\documentclass{article}

     \usepackage[preprint, nonatbib]{neurips_2019}

\usepackage[utf8]{inputenc} %
\usepackage[T1]{fontenc}    %
\usepackage{hyperref}       %
\usepackage{url}            %
\usepackage{booktabs}       %
\usepackage{amsfonts}       %
\usepackage{nicefrac}       %
\usepackage{microtype}      %
\usepackage{algorithm}
\usepackage{algorithmic}
\usepackage{hyperref}
\usepackage{biblatex}
\usepackage{graphicx}
\graphicspath{ {graphs/} }
\title{RBED : Reward Based Epsilon Decay}

\author{%
  Aakash Maroti\\
  \texttt{aakash.maroti@gmail.com} \\
}

\begin{document}

\maketitle

\begin{abstract}
  $\varepsilon$-greedy is a policy used to balance exploration and exploitation in many reinforcement learning setting. In cases where the agent uses some on-policy algorithm to learn optimal behaviour, it makes sense for the agent to explore more initially and eventually exploit more as it approaches the target behaviour. This shift from heavy exploration to heavy exploitation can be represented as decay in the $\varepsilon$ value, where $\varepsilon$ depicts the how much an agent is allowed to explore. This paper proposes a new approach to this $\varepsilon$ decay where the decay is based on feedback from the environment. This paper also compares and contrasts one such approach based on rewards and compares it against standard exponential decay. The new approach, in the environments tested,  produces more consistent results that on average perform better. 
\end{abstract}

\section{Introduction}

In Reinforcement Learning(RL) the core problem is often phrased as that of an agent
learning to interact with an environment. An agent's behaviour in an environment is
defined by its policy. We will be looking into a subset of such algorithms known as on-
policy algorithms. Here the agent has only one single policy governing its movement in
the environment. The agent uses the same policy to explore the environment and act in
it.

At any given point an agent can act greedily based on the information that it has to
maximize the total rewards that its expects to obtain. This is also known as exploitation.
Exploitation comes with a major drawback. The agent acts only on the basis of its
knowledge of the environment, which may be incomplete. This is especially problematic if
an agent has just began its interaction with the environment. Acting only greedily based
on it's limited knowledge of the environment makes it very unlikely for agent to learn the
optimal behaviour in the environment. An agent will never know about a different action and it's outcome in a given situation if it never tries it out. This is where exploration
comes in. When an agent explores it does not necessarily act to the best of its
knowledge, instead it explores different options available, dictated by some exploration
strategy. Thus, it is essential to strike a balance in between exploration and exploitation.

 \section{$\varepsilon$-greedy}

$\varepsilon$-greedy is one of the most popular strategies to balance exploration and exploitation. Here, with probability $\varepsilon$ the agent takes a random action, and with probability 1-$\varepsilon$ the agent takes the best action known to it\cite{Sutton1998}. It is from a class of $\varepsilon$-soft policies.

\subsection{The decay}

It is also important to note that exploration is more important when the agent does not have enough information about the environment its interacting with. Once an agent does have the information it needs to interact optimally with the environment, allowing it to exploit its knowledge makes more sense. We can thus see that the value of $\varepsilon$ should decay across the life of an agent to have it learn and act optimally eventually. A common way to obtain this by multiplying Epsilon by a real value less than 1 every episode. It is also known as exponential decay.

\subsection{The Feedback Based Decay}

The epsilon decay strategy that we have seen so far, follows a fixed mathematical curve,
that does not in any way take into account the performance of the agent in the
environment, or any feedback from the environment for that matter. This paper proposes an alternate solution, where rewards, feedback from the environment and agent's performance is taken into account when decaying the exploration rate. This can be challenging in cases where rewards accumulated by the agent, or the feedback it receives from the environment are not indicative of its performance. However, in cases where the environment provides proper goals and feedback, not only can the use of such approaches ease the reliance on expert
parameter tuning, they can often lead to better agents.

\subsection{Reward Based Decay}

This paper now suggests a form of feedback based decay, where the reward obtained from the environment is used to decay the exploration rate. Only when an agent has crossed some reward threshold, the value of $\varepsilon$ is reduced. Instead of assuming that the agent is learning more every episode, we wait for proof of agent's learning before reducing the epsilon value. Thus we set higher targets for agent every time $\varepsilon$ is lowered, and wait for the agent to reach the newer target before repeating the same steps again. Since this decay is not dependent on number of episodes, but the performance of the agent, it is possible for different agents with different learning capabilities to experience similar degree of exploitation based entirely on their performance.  

\subsection{Algorithm and Example}

We maintain a variable like reward threshold, over the course of decay which is initialized with a low value to begin with.
Here, we take a look at how a very simple reward based $\varepsilon$ decay would look like
\begin{algorithm}
\caption{$\varepsilon$-decay and reward threshold increment step}
\begin{algorithmic}
\IF{$last\_reward >= reward\_threshold$}
\STATE $\varepsilon \leftarrow decay(\varepsilon)$
\STATE $reward\_threshold \leftarrow increment(reward\_threshold)$
\ENDIF
\end{algorithmic}
\end{algorithm}

The decay and increment steps could be anything that fits the need of the problem. Even something as simple as adding and subtracting small fixed values work effectively for simple problems as openAI's cartpole-V0\cite{DBLP:journals/corr/BrockmanCPSSTZ16}. Above Algorithm is what would replace the simple $\varepsilon=\varepsilon*decay\_rate$ used in exponential decay. The key idea presented here is that only when the agent meets or surpasses the reward threshold, is it allowed to exploit more.

Let us work our way the hyper-parameters for OpenAi’s CartPole-v0. The environment is considered solved when the agent obtains an average score of 195+ in 100 consecutive episodes. The maximum attainable score from any episode is 200. A standard start and end value for $\varepsilon$ could be 1 and 0 respectively, where we start with maximum possible exploitation and end with maximum possible exploitation. Since all the rewards we obtain are whole numbers, we can at least increment target reward thresholds by 1 at the least to make newer thresholds more challenging than the last one. The final parameter that remains to be decided is the number of steps to take form maximum exploration to maximum exploitation. Since the problem statement has a well defined goal of reaching 195, it is possible to use that to decide the steps. We calculate the steps by assuming that the agent should reach the final goal threshold with exploitation alone. So, with this in mind, the number of steps can be set to the final target value.  All the values we have decided so far are fairly trivial and require almost no expert tuning. The decay step with these parameters is depicted in the algorithms below.

\begin{algorithm}
\caption{Initializing values for OpenAI's Cartpole-V0}
\begin{algorithmic}
\STATE $\varepsilon \leftarrow 1.0$
\STATE $min\_value \leftarrow 0.0$
\STATE $reward\_target \leftarrow 195$
\STATE $steps\_to\_take \leftarrow reward\_target$
\STATE $reward\_increment \leftarrow 1$
\STATE $reward\_threshold \leftarrow 0$
\STATE $change \leftarrow (\varepsilon - min_value )/steps\_to\_take$
\end{algorithmic}
\end{algorithm}

\begin{algorithm}
\caption{RBED for OpenAI's Cartpole-V0}
\begin{algorithmic}
\IF{$last\_reward >= reward\_threshold$}
\STATE $\varepsilon \leftarrow \varepsilon - change$
\STATE $reward\_threshold \leftarrow reward\_threshold + reward\_increment$
\ENDIF
\end{algorithmic}
\end{algorithm}

\section{Results}

RBED used in place of exponential decay, produces more stable results that are easier to replicate. Using the Neural Episodic Controller\cite{DBLP:journals/corr/PritzelUSBVHWB17} solution by Karpathy\cite{EpisodicController} showcased on the webpage for cartpole by OpenAI as a base, over 500\% improvement was seen in on the agent's ability to solve the environment within 500 episodes. Traditional exponential decay however, is earlier to reach the 200 points mark, and quicker to solving the environment whenever it does so. However it does so rarely, and we see a much more reliable performance using Reward-Based Epsilon Decay, as can be seen in figures 1 and 2. Using Reward-Based also outperforms exponential decay in every way when used with DQNs\cite{DBLP:journals/corr/MnihKSGAWR13}.Figure 1: Comparison of reward obtained by the agent at every given timestep, averaged across 20 runs. Reward based epsilon decay makes it more likely for the agent to cross the threshold of 195. The graph also depicts that exponential ‘peaks’ earlier when compared to rbed, however does not peak as high.

\begin{figure}
\minipage{0.32\textwidth}
  \includegraphics[width=\linewidth]{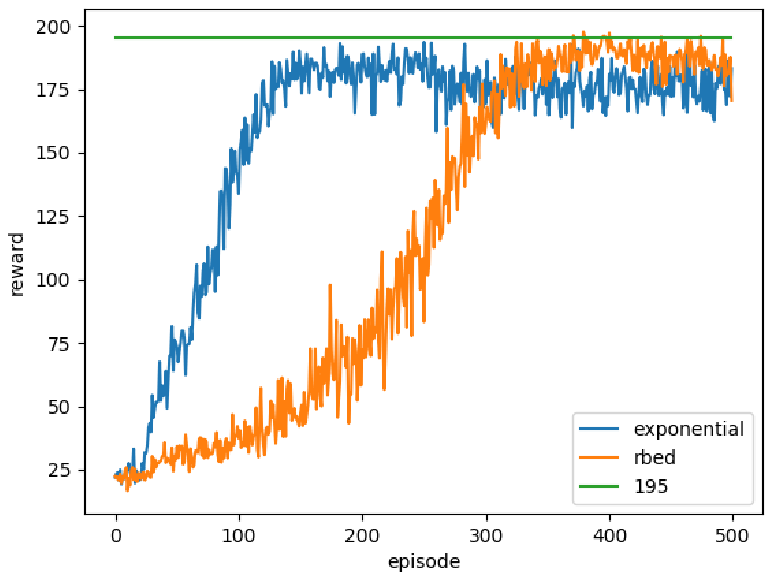}
  \caption{Comparison of reward obtained by the agent at every given episode, averaged across 20 runs. Rbed makes it more likely for the agent to cross the threshold of 195. The graph also depicts that while exponential ‘peaks’ earlier when compared to rbed, it fails to reach a value as high.}
\endminipage\hfill
\minipage{0.32\textwidth}
  \includegraphics[width=\linewidth]{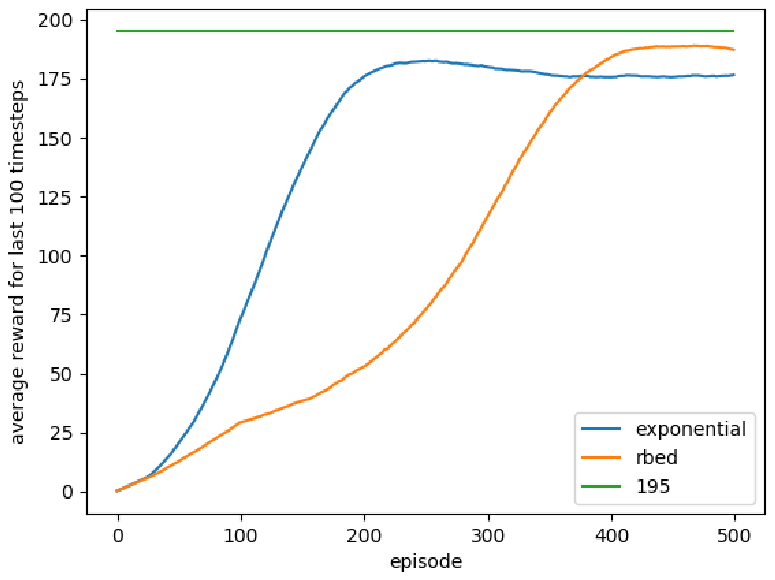}
  \caption{Comparison of average reward obtained across last 100 episodes, averaged across 20 runs. This is the metric used to determine the solution of the environment. The long term trend agrees with the experimental results that rbed is more successful in finding a solution.}
\endminipage\hfill
\minipage{0.32\textwidth}
  \includegraphics[width=\linewidth]{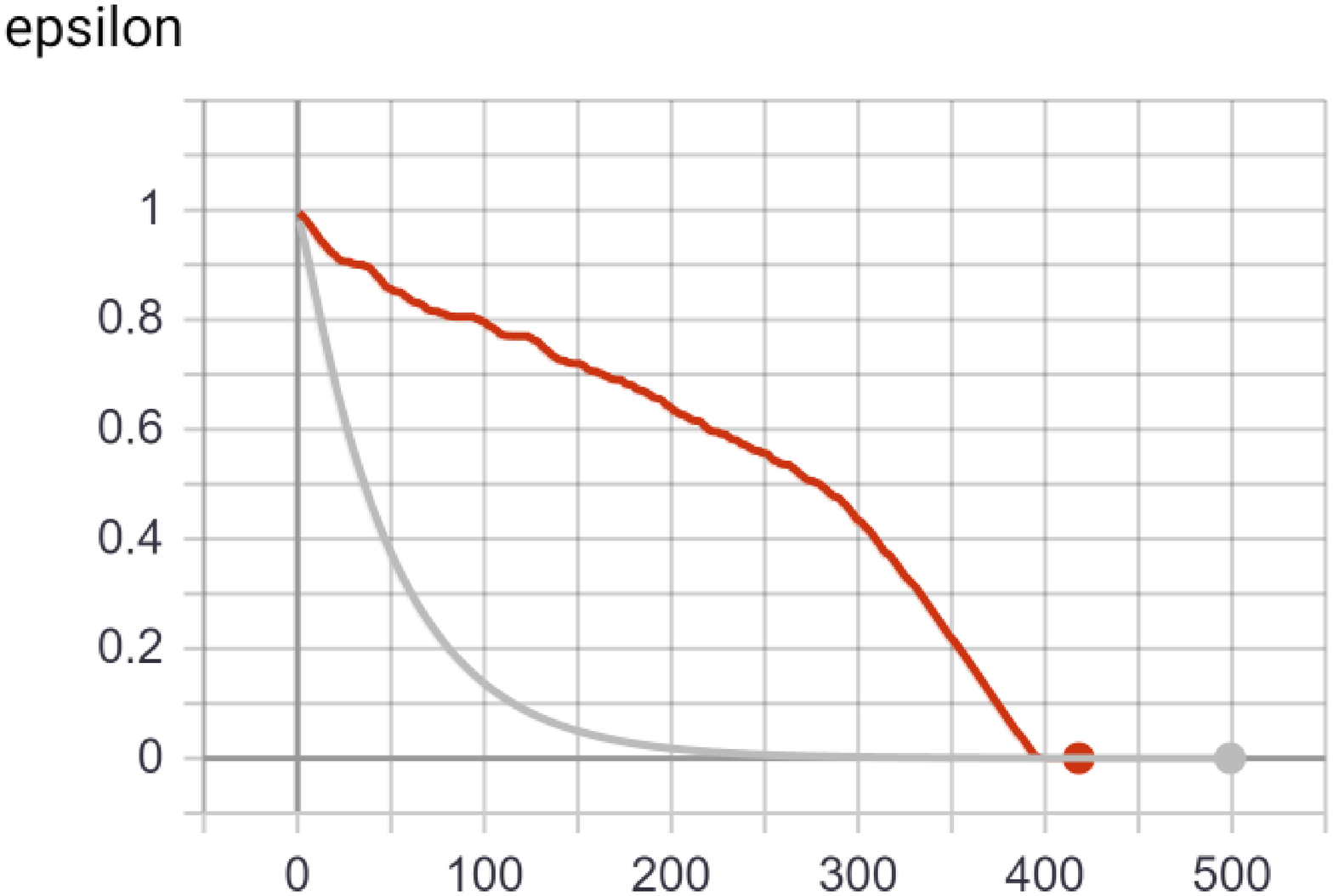}
  \caption{Comparison of how $\varepsilon$ decays across episodes. Red line represents Reward based decay and grey line represents exponential decay.}
\endminipage\hfill
\end{figure}

\section{Intuition}

Another way to interpret this decay strategy is based on responsibility. An agent is granted greater responsibility to act as it accomplishes greater rewards. Agent failing to meet the targets will be compelled to explore more until it does meet the targets.

\section{Conclusion}

While this decay strategy might not be suitable for a lot of environments and problem statements, it does bring a few good to have benefits in cases where it does. It removes a lot of uncertainty around the exploratory nature of an agent. Should the agent be exploring more? Has the agent explored enough? A reward based approach yields much easily understandable parameters which can be tweaked to get the desired results.
The parameters here are more related to the environment and task than they are to an agent. Agents, irrespective of their learning capabilities will get similar exploratory freedom at any given stage based on their accomplishments.

\printbibliography

\end{document}